\documentclass[12pt, peerreview]{IEEEtran}
\IEEEoverridecommandlockouts
\usepackage{cite}
\usepackage{amsmath,amssymb,amsfonts}
\usepackage{algorithmic}
\usepackage{graphicx}
\usepackage{textcomp}
\usepackage{xcolor}
\usepackage{dirtytalk}
\usepackage{multirow}
\usepackage{url}
\def\BibTeX{{\rm B\kern-.05em{\sc i\kern-.025em b}\kern-.08em
    T\kern-.1667em\lower.7ex\hbox{E}\kern-.125emX}}

\renewenvironment{thebibliography}[1]{%
  \begin{oldthebibliography}{#1}%
    \normalsize 
}{%
  \end{oldthebibliography}%
}

\begin{document}

\title{Advancing AI with Integrity: Ethical Challenges and Solutions in Neural Machine Translation\\

\thanks{}
}

\author{
\IEEEauthorblockN{Richard Kimera}
\IEEEauthorblockA{\textit{kimrichies@handong.ac.kr}} \\
\and
\IEEEauthorblockN{Yun-Seon Kim}
\IEEEauthorblockA{\textit{Sean0831@handong.edu} }\\

\and
\IEEEauthorblockN{Heeyoul Choi}
\IEEEauthorblockA{\textit{hchoi@handong.edu} }\\
}

\maketitle

\begin{abstract}
\normalsize
This paper addresses the ethical challenges of Artificial Intelligence in Neural Machine Translation (NMT) systems, emphasizing the imperative for developers to ensure fairness and cultural sensitivity. We investigate the ethical competence of AI models in NMT, examining the
Ethical considerations at each stage of NMT development including data handling, privacy, data ownership, and consent. We identify and address ethical issues through empirical studies. These include employing Transformer models for Luganda-English translations and enhancing efficiency with sentence mini-batching. And complementary studies that refine data labeling techniques and fine-tune BERT and Longformer models for analyzing Luganda and English social media content. Our second approach is a literature review from databases such as Google Scholar and platforms like GitHub. Additionally, the paper probes the distribution of responsibility between AI systems and humans, underscoring the essential role of human oversight in upholding NMT ethical standards. Incorporating a biblical perspective, we discuss the societal impact of NMT and the broader ethical responsibilities of developers, positing them as stewards accountable for the societal repercussions of their creations.
 \\

\end{abstract}


\section{Introduction}
Ethics, as an academic discipline, refers to the systematic inquiry and practical application of principles and values that serve as guiding frameworks for shaping human behavior and facilitating decision-making in matters of morality\cite{burget2017definitions}. It can be looked at in various perspectives and categories; Environment ethics, research ethics, ethics design methods that looks at human computer interaction, design research, professional ethics, medicine, and business, conflict of interest that focuses on the ability to distinguish between interest and moral ground, etc. The concept and perception can also vary according to cultures due to the various norms, definition of values, and beliefs. Ethics is crucial in multicultural societies to help in 1) promoting understanding and reducing conflict, 2) enhancing communication and cooperation, 3) addressing ethical dilemmas to enable diversity at workplaces, 4) maintaining social order, respect, and cohesion through guiding policy and practices\cite{josefova2016importance, natyavidushi2011importance, mccuen1991engineering}. 

The 1970s marked the emergence of computer ethics as a distinct discipline\cite{fenwick2022mapping}. The primary issues revolved around data privacy, security, and intellectual property rights. Information and Communication Technology (ICT) over the years, they expanded to encompass online privacy, intellectual property infringement, and the increasing prevalence of cyberbullying, prompting the formulation of specific laws and regulations\cite{Caccavale2022ToBF}. 

The advancing landscape of Artificial Intelligence (AI) systems necessitates a strong emphasis on ethics with responsible AI and digital ethics at the forefront, emphasising the need for transparency, accountability, and inclusivity\cite{Mirghaderi2023EthicsAT,Bandy2022InspectingAF}. As a sub-field of NLP, Machine translation (MT) has become very significant in today society due to the multilingual communities. MT tools can help scientists take concrete steps toward publishing in multiple languages, including in English, and can be harnessed to improve online, international global health education, facilitating more equitable and inclusive online learning environments\cite{hill2022lost}.  Building of such systems involves the use of algorithms to receive a sequence of text from one language and transforming it into its corresponding sequence in another language. These systems need to comply to the rules and procedures that are acceptable in community in both hugh and low resource communities.   
 
MT inherently mirrors the values of its developers in both the tools used for interacting with it and the resulting translated text. In alignment with the broader context of technological advancements, it is essential to recognize that neither the development process nor the final output of MT can be regarded as neutral\cite{moorkens2022ethics}. Instead, they invariably propagate the viewpoint and perspective of the developers or the translators who contributed to the training data. Consequently, it is paramount to deliberate upon the ethical implications associated with MT's development and its produced output. Mindful and deliberate decisions must be taken when instituting a data-driven MT workflow, considering the inherent biases and perspectives that might be embedded within. The use of MT with low-resourced languages (LRL's) can further raise concerns that can lead to the marginalization of underrepresented perspectives and communities\cite{Liu2022SustainabilityAI, mager2023ethical, Haroutunian2022EthicalCF}. 
 
In this paper therefore, we provide an outline of general ethical concerns in NLP, and gradually narrow our focus to Neural Machine Translation (NMT). This approach helps elucidate specific ethical dilemmas faced by researchers in this subfield. Following this, we delve into the complex question of whether machines can possess ethical considerations and, subsequently, who bears responsibility in instances of unethical machine behavior. This is a particularly pressing issue in the realm of autonomous decision-making systems such as Generative AI systems.

A distinctive feature of our analysis is the incorporation of perspectives from the Bible. We examine how these ancient texts can contribute to the modern discourse on AI ethics, offering a broader, more holistic view of the ethical landscape in technology. This cross-disciplinary approach not only enriches the ethical discussion in AI but also bridges the gap between technological and philosophical realms.

This paper, therefore, adopts a comprehensive perspective and seeks to articulate the nuanced ethical intricacies specific to NMT. Addressing these concerns goes beyond academic inquiry, recognizing that the decisions made in developing and deploying NMT systems carry significant consequences. This necessitates exploring the ethical capacity of AI-driven translation models and the moral responsibility attributed to their creators and operators. Incorporating a biblical perspective, we approach these modern ethical challenges with an understanding that echoes timeless virtues, such as wisdom, integrity, and the pursuit of communal well-being, akin to the ethical teachings in the bible. This cross-disciplinary approach enriches the ethical discourse in AI, bridging the gap between rapid technological advancement and the need for conscientious stewardship over our creations. Therefore, this paper frames pressing questions within this broader ethical context: Can AI systems reflect the ethical deliberation akin to these virtues in their functionality, and how do we assign responsibility for their actions? These queries become increasingly critical as autonomous decision-making systems gain prevalence.
The paper is positioned within this framework, converging on the objective to scrutinize ethical considerations pertinent to NMT. We endeavor to reveal the ethical implications at every development stage of these translation systems and propose guiding principles that balance technical progress with ethical accountability. Through this inquiry, we aim to foster the advancement of NMT technologies that honor the diversity of human languages and cultures, promoting justice and equity for communities of all linguistic resources.

\section{Literature Review}
Neural Machine Translation (NMT) is a form of machine translation that utilizes artificial neural networks to predict the likelihood of a sequence of words, typically modeling entire sentences in a single integrated model. NMT aims to translate whole sentences at a time, rather than piece by piece, which allows for more fluent and accurate outputs by considering the context of the input sequence \cite{Jain2021LiteratureSN}. Nonetheless, the necessity for substantial volumes of parallel data is a pivotal factor influencing the effectiveness of NMT models \cite{UlQumar2023NeuralMT}. This poses a formidable obstacle, particularly for low-resource languages where parallel corpora are in short supply. In this review, we will center our attention on the ethical dimensions encompassing NMT, for both high-resource and low-resource languages.

\subsection{Ethical Integration in AI: Frameworks and Principles}
Ethical frameworks provide a structure for ensuring that technological innovations adhere to moral principles and societal values. The Menlo Report, inspired by the Belmont Report, adapts its principles to the ICT field, particularly in cybersecurity, emphasizing respect for persons, beneficence, justice, and respect for law and public interests \cite{bailey2012menlo}. In technology, some researchers have utilized the Value Sensitive Design (VSD) \cite{winkler2021twenty} to integrate human values into the design of ethical AI \cite{umbrello2021mapping}, aiming to mitigate ethical concerns. VSD consists of conceptual, empirical, and technical investigations. Additionally, the Ethical Risk Assessment (ERA) framework has been applied to systems like robotics to minimize ethical risks \cite{winfield2020roboted}, while the European Commission has introduced the Responsible Research and Innovation (RRI) framework, highlighting co-creation and co-production \cite{Pansera2020EmbeddingRI}.

The Ethical Impact Assessment (EIA) is another framework that evaluates the potential ethical impacts of new technologies, considering factors like data privacy, user consent, potential for harm, and societal impacts \cite{Reijers2017ACF}. Moreover, the IEEE Ethically Aligned Design Framework ensures that autonomous and intelligent systems prioritize ethics, focusing on privacy, transparency, human well-being, trust, and accountability \cite{Ema2018DesignDA}. However, it, along with other frameworks mentioned, has been critiqued for lacking specificity \cite{Kazim2020HumanCA}.

While these frameworks offer foundational principles and methodologies for the ethical development of technology, the ever-evolving nature of technology presents complexity and dynamism, requiring continuous refinement and contextual adaptation. Acknowledging this, our assessment has not adhered strictly to any single framework, as none fully addressed our ethical concerns comprehensively. Instead, we have focused on general principles pertinent to NMT ethics, such as algorithmic bias, fairness, explainability, and machine autonomy.

Future research should aim to develop more tailored ethical guidelines for various AI and computing domains, ensuring that ethical considerations keep pace with rapid technological progress and the nuances of specific research areas.

Ethical guidelines in NMT are particularly critical at different stages of model development. For example, pre-training or fine-tuning models on extensive datasets can introduce inductive social biases, affecting translation fairness \cite{meade2022empirical}. Moreover, NMT models often struggle with gender biases, which can negatively impact users and society. As NMT technology evolves, it is imperative to adopt a comprehensive approach to ethical considerations, encompassing not only technological interventions but also promoting a culture of openness, responsibility, and inclusive dialogue among stakeholders.

We strive to demonstrate how these challenges can be addressed so that NMT systems are utilized equitably and respectfully, mindful of their diverse societal impacts. In doing so, NMT can realize its potential as a tool for bridging linguistic and cultural divides, fostering global communication while firmly grounded in ethical integrity.

\section{Methodology}
In this scholarly pursuit, the authors embraced a two-fold methodological approach to address ethical issues within NMT. First, the investigation followed the  aurthors empirical studies, of four pivotal works. These included one that aimed at the creation of a corpus and the utilisation of the transformer model for training NMT models between Luganda and English \cite{kimera2023building}, an efficient approach in expediting the training process using mini-batching techniques \cite{rim2023mini}, enhancing the data labeling techniques and finetuning of the longformer moder to classing depression severity using text from Reddit \cite{kimera2023enhanced}, and fine-tuning BERT for social media data in both Luganda and English \cite{kimera2023fine}.

To further substantiate the findings, the authors conducted a literature search, extracting relevant studies from four academic databases—Google Scholar, Semantic Scholar, Nature, and the arXiv repository. This review was complemented by selected grey literature deemed credible within the deep learning community, including repositories such as GitHub, Kaggle, and Hugging Face. This literature survey was meticulously curated to focus on the ethics of machine translation, encompassing works related to both low-resource and high-resource languages, while deliberately excluding non-relevant government agency literature and publications that did not specifically address machine learning, deep learning, or machine translation in any language resource context. Through this academic exercise, the authors aimed to validate the ethical solutions proposed, ensuring a robust foundation for the contributions to the field of NMT.

\section{Results: Ethical Issues and solutions }
In this section, the authors discuss ethical concerns encountered or observed during their research process. They detail these challenges, describe the strategies they have employed to address these ethical dilemmas and incorporate insights and methodologies from existing literature. 

The first section presents concerns that are Human controllable, where as a second section attempts to answer whether AI models have ethics. 

\subsection{Data Collection and preparation}

\subsubsection{Violating data sharing and usage Licenses}
In our exploration of NMT, the journey led us to diverse datasets hosted on platforms like GitHub and Hugging Face. Delving into these repositories, we encountered a myriad of bilingual and multilingual datasets, each presenting an opportunity to refine our models for English and Luganda. The process extended beyond mere data extraction; through web scraping, we engaged deeply with the datasets, employing n-gram analysis and topic modeling to glean contextual insights without directly incorporating the data into our models.
            
Faced with the imperative of ethical compliance, we meticulously reviewed the licensing terms of each dataset, including Creative Commons and the MIT license. This examination ensured that our utilization of these resources was not only legally sound but also aligned with the broader ethos of academic integrity. By giving proper accreditation and adhering to the stipulated conditions, we navigated the ethical landscape, recognizing the importance of understanding and respecting the licensing agreements that underpin dataset usage in academia.
            
\subsubsection{Copyrights on Data Tools.}
\cite{kimera2023enhanced} presented the researchers with a challenge when selecting the appropriate tools for the study. The use of the Patient Health Questionnaire-9 (PHQ-9) initially seemed promising, given its widespread availability online and its potential for extracting keywords for pattern matching in dataset labeling. However, the realization that PHQ-9 is copyrighted by Pfizer Corporation introduced an ethical dilemma regarding the permissible use of copyrighted materials in academic research. In response to this issue, our team transitioned to employing Beck's Depression Inventory\cite{jackson2016beck}. This tool, while also copyrighted, allowed for its use within the parameters of our study without infringing on copyright terms, a decision supported by consultations with medical experts who endorsed its application. This experience underscored the importance of diligently seeking the consent of copyright holders, highlighting the necessity for researchers to engage with the creators or legal representatives of such tools to secure the appropriate licenses for use.

\subsubsection{Accreditation and pre-requisites: The MIMIC Dataset Example} 

In the realm of healthcare research, the temptation to leverage readily available secondary data can sometimes lead to overlooked ethical considerations. This was notably evident in our attempt to utilize the MIMIC dataset, a comprehensive healthcare dataset with availability on platforms like Kaggle. The challenge arose when the need to extract and translate specific data components potentially conflicted with the original data usage policies set forth by the publishers. To address this challenge and ensure compliance with ethical standards, our approach emphasized direct engagement with the original sources of the dataset. By accessing MIMIC directly from the Physionet website \cite{johnson2023mimic}, we adhered to a rigorous protocol that included signing a Data Usage Agreement, abiding by the PhysioNet Credentialed Health Data License, and completing a CITI Data or Specimens Only Research Course. This methodical approach not only safeguarded against inadvertent policy violations but also reinforced our commitment to ethical research practices within the sensitive domain of healthcare data.
            
\subsubsection{User privacy.}
The authors faced an ethical dilemma in retrieving data from social media platforms, where user-identifiable information like usernames and opinions is often accessible through APIs without explicit user consent. This challenge was particularly pronounced with data collection efforts on Twitter and Reddit, platforms known for their vast user-generated content and the presence of anonymous users or bots, making it impractical to seek individual consent.

In navigating this ethical landscape, the authors adopted a rigorous approach to data retrieval. By exclusively focusing on the extraction of textual content and meticulously cleansing the data of any potentially identifiable information, they ensured that the integrity of user privacy was upheld. This conscientious process resulted in datasets devoid of identifiable markers, comprising solely non-identifiable plain text. Although this method limited the range of data analysis techniques available due to the absence of certain variables, it underscored the authors' commitment to privacy. 
            
Moreover, adherence to the terms of use specified by the APIs of these platforms was paramount. The authors rigorously complied with these stipulations, which encompassed prohibitions on reverse engineering, restrictions on commercial use, and mandates for the secure handling of data and credentials. This adherence not only ensured the ethical integrity of their research but also highlighted the importance of operating within the legal frameworks established by social media entities.
            
\subsubsection{Data credibility and representation.} 

The foundation of NMT models' fairness and representativeness heavily relies on the source and diversity of the data employed. Such diversity is paramount not only for earning the research community's trust and acceptance but also for the resultant quality of the models' outputs. This is particularly true for low-resource languages, where data scarcity poses a significant challenge. Often, the data that is accessible, such as biblical texts, may not be suitable for various tasks due to its narrow scope and limited generalizability, despite being available in numerous languages. This situation underscores the critical need for datasets that are both varied and contextually relevant, especially in the context of low-resource languages. 

To address this issue, researchers are tasked with ensuring the diversity of their data sources. This can be achieved by examining the data collection methods of dataset publishers or by establishing clear data collection and quality assurance procedures. In scenarios where data is scarce, innovative techniques like back translation and transfer learning offer viable solutions. Additionally, data augmentation methods, including Easy Distributional Data Augmentation (EDDA), Type Specific Similar word Replacement (TSSR) \cite{mahamud2023distributional}, and Subtree swapping \cite{dehouck2020data}, can be instrumental in enriching datasets. These strategies collectively contribute to the development of NMT models that are not only fair and representative but also capable of producing high-quality outputs across a broad spectrum of languages, thereby enhancing the models' utility and applicability in real-world scenarios.   
            
\subsubsection{Safeguarding Data Privacy in Remote Collaboration and Research}
In the era of remote working and multidisciplinary team collaboration, the convenience of online tools for data sharing has become indispensable for research activities like translation and annotation. Yet, this ease of data exchange, exemplified by the simplicity of transferring information via email, harbors potential risks to user data privacy if meticulous data protection measures are not rigorously applied. Such scenarios underscore the imperative for robust data security protocols in collaborative research settings to prevent inadvertent data breaches.
            
To mitigate these risks, it is essential to establish comprehensive safety guidelines encompassing data sharing, access, and storage. Adopting industry-accepted standards, such as two-factor authentication, de-identification, anonymization, data encryption, and data minimization, can significantly enhance the security of shared data. Furthermore, strict adherence to prevailing regulations, including the Health Insurance Portability and Accountability Act (HIPAA) and the General Data Protection Regulation (GDPR), is critical. These practices collectively form a robust framework for protecting user data, ensuring that research collaboration in the digital realm remains both productive and secure.  
            
\subsubsection{Addressing bias in translation datasets.}
The lack of a comprehensive strategy to counteract bias in translation datasets emerges as a pivotal ethical concern, especially when it leads to the generation of content that is discriminatory, offensive, inaccurate, or incomplete. Such biases not only perpetuate harmful stereotypes but also undermine the integrity and quality of translations, posing significant ethical and reputational risks. 
            
To combat these challenges, it is crucial to establish clear guidelines aimed at minimizing bias within datasets. This involves employing human reviewers to assess the accuracy, cultural sensitivity, and fairness of the data. Offensive or discriminatory language can be mitigated through techniques such as filtering, anonymizing, or modifying data points to enhance the overall quality of the dataset. Furthermore, employing debiasing methods, including contrastive learning between clean and noisy sentence pairs \cite{huang2023improving}, presents a proactive approach to refining dataset integrity. Although reviewing each sentence individually may be labor-intensive, it is a highly recommended practice to ensure thoroughness in eliminating bias. These measures collectively contribute to fostering ethically responsible and high-quality translation outputs, safeguarding against the adverse effects of dataset bias. 

\subsubsection{Cultural representation.}

The NMT tools, while economically appealing and sometimes the only option in the absence of human translators could lead to potential loss of semantics and context inherent to the source language, which can substantially diminish the accuracy and integrity of the translated content.

To address these concerns, implementing human post-editing or engaging translation teams proficient in both the source and target languages \cite{Gurov2022LEARNINGNM} emerges as an effective strategy. Additionally, maintaining transparency regarding the translation methodologies employed can enhance the credibility and reliability of the translated data. Researchers are encouraged to adhere to a systematic translation process that encompasses stages such as preparation, translation, pretesting, revision, and documentation \cite{Gurov2022LEARNINGNM}. When dealing with low-resource languages, such as Luganda, it is imperative to involve and consult all relevant stakeholders throughout the deployment of NMT systems, ensuring that ethical considerations and quality standards are rigorously upheld.

\subsection{data labelling} 

\subsubsection{AI assisted data labelling.}

The advent of large language models (LLMs) has revolutionized data labeling, offering models like BART\cite{kimera2023enhanced} that facilitate annotation without human expertise. These models employ natural language inferencing to evaluate premises and hypotheses, assigning probability scores to determine the similarity across various hypotheses. Such capabilities enable the annotation of data in multiple languages with unprecedented ease. However, this scenario raises ethical concerns regarding the accuracy and potential bias of AI-generated annotations, which could compromise the quality and fairness of the labeled data.

To mitigate these risks, integrating human oversight into the AI labeling process is crucial. Expert annotators play a pivotal role in reviewing and validating AI-generated annotations, ensuring the data's quality and correctness. Additionally, fine-tuning AI models to align more closely with the specific requirements of the data labeling task can significantly enhance annotation quality. Custom training regimes tailored to the unique characteristics of the dataset can further refine the accuracy of AI-generated labels. In situations where bias and inaccuracy are concerns, employing explainable AI \cite{rubinic2023artificial} offers an additional layer of transparency and understanding, allowing researchers to better grasp and control the labeling process. This holistic approach to AI-assisted data labeling underscores the importance of balancing technological advancements with ethical diligence to uphold the integrity of research data.

\subsubsection{Balancing expertise and ethics in human annotated data}

Employing human annotators for data labeling, especially within specialized domains, raises ethical challenges related to the annotators' level of expertise and the tools at their disposal. The reliance on open datasets, already labeled for specific applications, further complicates these ethical considerations. A significant ethical concern is the introduction of human bias into the labeled data, stemming from annotators' personal biases, varying levels of expertise, or the absence of explicit annotation guidelines.

To counteract these challenges, engaging a diverse and representative cohort of annotators is crucial. Achieving diversity involves recruiting annotators from varied backgrounds and expertise levels, all the while equipping them with comprehensive annotation guidelines and conducting regular quality assessments. In an experimental setup\cite{kimera2023enhanced}, the blend of AI models with human expertise for data labeling, followed by majority voting to determine the main label, exemplifies a practical approach to minimizing bias.

Moreover, the ethical dimension of developing resources for low-resource languages, such as Luganda, necessitates community involvement. Collaborating with native speakers, linguists, and cultural experts not only ensures the accurate representation of the language and culture but also aligns the development of language processing tools with ethical standards. This collaborative ethos fosters the creation of language processing solutions that are not only effective but also ethically grounded and culturally sensitive.
            
\subsubsection{Conflict of interest.}

The integrity of data labeling can be compromised by non-disclosure of conflicts of interest (CoI), which encompass financial interests, affiliations, incentives, or personal relationships potentially influencing research outcomes. 

To safeguard against these risks, disclosing any potential conflicts of interest is paramount, alongside implementing strategies to mitigate their impact through well-defined policies and guidelines. An independent review of the labeled dataset serves as a critical mechanism for detecting any biases that may stem from undisclosed CoIs. Additionally, equipping researchers with the necessary training to recognize CoIs and fostering ethical collaboration practices, particularly in domain-specific research, are essential steps in maintaining the integrity of the research process \cite{Kooli2022ArtificialII}. This comprehensive approach ensures that data labeling is conducted in an environment that prioritizes ethical considerations, transparency, and objectivity.

\subsubsection{Preserving Context }

The text cleaning phase presents an ethical quandary, especially when elements crucial for maintaining sentence context, such as numbers or social media peculiarities (jargons, emojis, emoticons), are removed or altered. Consider the significant shift in meaning when '1200' is extracted from a sentence like 'I am 1200kms away from the destination.' Such modifications can compromise the natural structure of language and inadvertently reduce the dataset's comprehensiveness, impacting the training and effectiveness of Machine Translation models.

A refined approach to data cleaning, particularly for social media-derived datasets, is critical. It entails a context-sensitive review of each sentence, ensuring numbers or symbols are thoughtfully replaced with text that preserves the original context. Leveraging libraries and APIs to translate emojis and emoticons into textual descriptions is a strategy that can prevent the loss of meaning\cite{czestochowska2022contextfree}. Adopting these meticulous data cleaning practices is vital for upholding the linguistic integrity of datasets, thus enhancing the performance and reliability of Machine Translation models.
        
\subsection{Model Training} 

\subsubsection{Tackling Imbalances in NMT.}

The challenge of using imbalanced datasets in classification tasks within machine learning, particularly for language-specific data, raises ethical concerns regarding model performance. An imbalanced dataset may compromise the integrity of a model, embedding biases and resulting in predictions that are both unfair and inaccurate\cite{kimera2023enhanced}.

To counteract these issues, ensemble learning methods emerge as a viable solution, leveraging the collective strength of multiple models to enhance classification performance\cite{Gholami2023OnEO}. Data augmentation serves as an additional strategy to balance datasets. While applying these techniques to language-specific data might present complexities, exploring options such as paraphrasing, synonym substitution, or language-specific transformations can effectively enrich the dataset. It is crucial for researchers to consider the nuances and contexts of the target language during augmentation to preserve the original meaning.

Continuous evaluation of the model's performance is paramount, with a particular focus on identifying and mitigating bias and inaccuracies in classifying underrepresented groups. Employing metrics such as F1 score, recall, precision, accuracy, BLEU score, perplexity, and CHRF score can provide comprehensive insights into the model's effectiveness and fairness\cite{chauhan2022comprehensive}. These measures ensure that the model's training process remains aligned with ethical standards, promoting fairness and accuracy in NMT applications.
            
\subsubsection{Using open source codes.}

The practice of leveraging online code resources, from snippets to comprehensive projects, is commonplace in software development and NMT research. Yet, the importance of ascertaining the quality and reliability of such code cannot be overstated, as substandard code can adversely affect research outcomes and software integrity. Addressing the ethical considerations associated with using open-source code necessitates a thorough examination of code quality. Tools like SonarQube and CodeClimate offer valuable insights by identifying code smells and potential bugs\cite{widder2022limits}. Furthermore, implementing additional testing measures, such as automating tests with Selenium and JUnit, enhances the code's reliability. Conducting team code reviews and assessing the author's credibility through platforms like GitHub and LinkedIn are also crucial steps in ensuring code integrity.

Security assessments are equally important, with tools like OWASP ZAP and Burp Suite serving as instrumental in uncovering securit vulnerabilities\cite{lysaght2020watching}. Additionally, SPDX can be utilized to clarify the licensing status of the code, ensuring compliance with legal and ethical standards\cite{zasiekin2020ethical}. Together, these practices form a comprehensive approach to mitigating the risks associated with using online code resources, ensuring that NMT research and software development are grounded in quality, reliability, and ethical standards.  
            
\subsubsection{Optimizing NMT with appropriate tokenization methods}

In the development of Neural Machine Translation (NMT) models, tokenization plays a pivotal role in deconstructing text into manageable units, such as words, sentences, or subwords. The selection of a tokenization method is crucial, influenced by the developer's preference, the language characteristics, and the model's architecture. For the authors, finding a method adept at processing both analytical languages like English, with minimal inflection, and morphologically rich languages like Luganda was imperative. Inadequate tokenization could introduce bias, compromise translation quality, and disregard cultural and linguistic nuances.
            
To address these challenges, the authors employed Byte Pair Encoding (BPE)\cite{kimera2023building}, a subword tokenization technique recognized for its effectiveness across diverse language structures. BPE adeptly maintains the linguistic integrity of both English and Luganda by leveraging the compositionality of sub-words, offering a nuanced approach that respects the morphological complexity of each language. Additionally, this method provides a robust solution for managing out-of-vocabulary (OOV) words\cite{provilkov2020bpedropout}, further enhancing the model's capacity to produce accurate and culturally sensitive translations. This strategic choice in tokenization underscores the authors' commitment to quality and inclusivity in NMT model training.

\subsection{Publications}

\subsubsection{Dataset Publication}

The proliferation of internet public spaces has significantly enhanced researchers' ability to disseminate their datasets across diverse platforms, including GitHub, Kaggle, and Hugging Face. This openness facilitates unprecedented collaboration and innovation within the realms of code-sharing, data science, and machine learning. Yet, the practice is not devoid of ethical complexities, notably concerning privacy, data ownership, and the imperative of de-identification. These concerns are magnified in cases where user consent for data usage is absent or when dataset licenses are inadvertently breached, highlighting the paramount importance of adhering to ethical standards in research data sharing.

In navigating these ethical waters, a profound comprehension of the guidelines and legal stipulations governing data sharing and publication becomes essential\cite{cole2021ten}. As of this writing, the authors have exercised caution, choosing not to release the datasets utilized in the health related research\cite{kimera2023enhanced}. This decision stems from a conscientious evaluation of the ethical implications, particularly given that while the bulk of the data was sourced from the public domain, segments obtained from social media and healthcare domains warrant additional scrutiny. Moving forward, the authors anticipate responsibly sharing their datasets, ensuring compliance with ethical norms and legal requirements, thereby contributing to the collective knowledge base in an ethical and legally sound manner.

\subsubsection{Publication of Pretrained Models}
The dissemination and publication of pretrained models in the public domain is pivotal. It serves the purpose of advancing open access to these models and resources, with a particular emphasis on making them available for languages such as Luganda. Such an initiative can greatly enhance the accessibility of translation capabilities and ultimately prove advantageous to a wider audience. However, one should be careful about research/innovations ownership by checking with universities ethical guidelines on research and innovation.

\subsubsection{Submission to Jounals}
The pressures of academic timelines may tempt researchers to submit their work to multiple journals simultaneously, aiming to accelerate the publication process. However, this approach, while seemingly efficient, poses ethical challenges and risks breaching the submission policies of academic journals. Scholars are encouraged to uphold the integrity of scholarly publishing by avoiding concurrent submissions. Adhering to these ethical guidelines not only maintains the respectability of the academic community but also prompts the exploration of alternative strategies for managing publication timelines within ethical bounds.

\subsubsection{Sharing pre-prints post publication}

Posting published work on pre-print servers is beneficial for garnering community feedback. Yet, post-publication, it's critical to align with journal guidelines and copyright agreements, either by updating or removing the pre-print. This adherence safeguards ethical and legal standards in academic dissemination.

\subsubsection{Upholding Transparency and Reproducibility}
It is incumbent upon the researcher to ensure that the research methods, data, and findings they present are characterized by transparency and reproducibility. This commitment to transparency is vital to uphold the integrity of the research process. Violating these ethical standards can impede the growth and progress of research, potentially eroding trust within the academic and scientific community. Therefore, it is imperative to prioritize transparency and reproducibility in one's research endeavors to maintain the highest standards of integrity.

\subsubsection{Acknowledgment Practices in Research Collaborations and Tool Usage}

Researchers are encouraged to acknowledge the contributions of their collaborators, a key practice for ensuring equitable authorship and upholding collaboration standards. The debate around recognizing contributions of text generation tools, such as ChatGPT, depends on their application. For non-native speakers, these tools offer significant aid in grammar checking and translation. However, careful scrutiny for plagiarism is essential before journal submissions to preserve academic integrity and the ethical standards of research publications.

\section{Do AI models have Ethics?}
Recent developments in LLM's, particularly in Natural Language Generation, have brought to light significant ethical considerations in the field of NMT. As  \cite{zasiekin2020ethical} argues that translation quality is not just a technical issue but also an ethical one, encompassing aspects of the translation as a product, a process, and an industry. 

Therefore, it is imperative to assess Neural Machine Translation (NMT) not only from a human standpoint but also from the perspective of the machine/product. The authors acknowledge the ongoing debate regarding the attribution of ethical responsibility to machines. They present this not as a definitive stance but rather as a series of discussion points that require continuous evaluation, especially given the rapid advancements in the field. This approach underscores the importance of re-examining and evolving our understanding of ethical responsibility in the context of machine intelligence and its applications

\subsection{The Blurred Lines of Ethical Responsibility}
At a basic level, machines lack inherent ethics as they are not directly responsible for actions unless initiated by human beings. However, the unpredictability of text generated by modern NMT systems complicates this view. These advanced models are capable of creating content that extends beyond the scope of their training data, leading to unique and previously unthought-of ideas. In the context of NMT systems, humans are responsible for a significant portion of these concerns, arguably up to 80\%. Nevertheless, when an NMT system operates independently, without human oversight, the question of ethical responsibility for the output becomes more complex. 

\subsection{The Role of Language Embeddings in NMT}
In the realm of Neural Machine Translation (NMT), language embeddings play a critical role, representing words or phrases within a continuous vector space and enabling the model to discern relationships between words in different languages\cite{mikolov2013distributed}. However, the variability in these embeddings, influenced by factors like training conditions and model architecture, raises a fundamental ethical question: in cases of translation inaccuracies or biases, should the blame be attributed to the machine, or to the humans who designed and trained it?

This question becomes particularly poignant considering the attention mechanism in NMT, which is susceptible to the length of text per sentence and impacts the translation's accuracy and quality. If the model's 'attention' leads to a skewed translation, is it a flaw in the machine's learning process, or is it a reflection of the dataset and parameters set by human developers? The machine, after all, operates within the confines of its programming and training, both of which are human-controlled aspects.

Moreover, while machine resources might affect the training scale, they do not typically alter embeddings during inference for a fixed model. This again shifts the focus to human decisions made during the model's development phase. Thus, even though the machine executes the translation, the ethical responsibility for its output arguably lies more with the humans who develop, train, and deploy the NMT system.  

This perspective suggests that while machines can "make decisions" based on their programming, the ultimate ethical accountability rests with the humans who create and manage these systems. Therefore, in the context of NMT, any ethical blame for translation errors or biases should be directed more towards the developers rather than the machine itself, emphasizing the need for responsible AI development practices. 

\subsection{The Imperative for Human Supervision}
The absence of human supervision in NMT can lead to translations that are offensive, harmful, or not aligned with community standards. Such incidents can cause significant distress, perpetuate misinformation, or even harm international relations. Ensuring that the outputs of NMT systems adhere to societal norms and standards is a critical aspect of ethical responsibility in this field. This can be achieved through content filters, guidelines, and regular quality checks.

\subsection{Decision making ability}
Current AI systems, including neural networks, operate based on algorithms and data. They do not possess consciousness or moral understanding, in fact LLMs like chatGPT cannot be relied on to provide ethical guidance since there outputs can vary if the same query is passed more than once, take an example of the trolley problem and its pro-environment nature and left-wing liberal ideology\cite{krugel2023moral}.  

The fundamental difference between neural networks and human moral decision-making lies in consciousness and intent. Humans make ethical decisions based on a complex interplay of consciousness, cultural background, personal experiences, and moral education. Neural networks, on the other hand, rely solely on mathematical functions, learned patterns, and the data they have been trained on. They lack the ability to consciously reflect or understand the moral weight of their "decisions". 

The discussion of whether machines have ethics or not will continue to create a large debate. Looking at how machines learn and base there decisions limited by data availability is not far from how humans learn as they too equally have limited knowledge. Humans learn through experiences and hence keep updating their knowledge, a concept that machines equally if calibrated.

\section{Discussion}

In this exploration, the authors navigate the ethical landscape within a technical context, addressing significant issues and proposing targeted solutions. However, this inquiry presents only a segment of the wider academic journey. Extending this exploration reveals the profound impact of ethical standards on one's professional pathway, especially within the mentor-student dynamic integral to academia's educational, research, and teaching frameworks \cite{folse1991ethics}. Incorporating a spiritual dimension offers a fuller perspective. For instance, the encouragement found in "Whatever you do, work heartily, as for the Lord and not for men" (Colossians 3:23-24) can motivate developers to pursue excellence and maintain the highest professional standards. Similarly, "Do you see a man skilled in his work? He will stand before kings; he will not stand before obscure men" (Proverbs 22:29) recognizes the importance of skill and diligence in career progression. Moreover, the emphasis on integrity and honesty as essential for ethical conduct is captured in "The integrity of the upright guides them" (Proverbs 11:3), while the value of diligence as a key to success is articulated in "The soul of the sluggard craves and gets nothing, while the soul of the diligent is richly supplied" (Proverbs 13:4). These Biblical passages underline the significance of a solid ethical foundation in academic and professional endeavors, highlighting virtues such as integrity, excellence, and diligence as critical for shaping one's professional identity and practices\cite{folse1991ethics}. Embracing these values during graduate education is vital for career development and professional achievement, illustrating the profound relevance of our research to the academic community.

Academia's 'publish or perish' ethos, often pressures academicians to sideline ethical norms in their quest to feature in high-impact journals. To counter this, the development of forums and training programs is imperative, not only to enhance ethical understanding among academicians but also to foster a culture of continuous professional growth and knowledge exchange \cite{churchill1982teaching}.

The exploration of ethical concerns and proposed solutions in the context of Neural Machine Translation (NMT) development underscores the crucial role of self-education for developers. This is paralleled by the Biblical emphasis on the pursuit of wisdom and understanding, as encapsulated in Proverbs 4:7: "Wisdom is the principal thing; therefore get wisdom: and with all thy getting get understanding." Developers are urged to diligently understand website policies, rules, and guidelines, mirroring the Biblical exhortation to seek knowledge and advice (Proverbs 19:20). Ignoring established standards, such as those set by the authors of datasets, carries significant implications \cite{jefferson2022graimatter}, reflecting the caution against neglecting wisdom in Proverbs 1:29-31. Such oversight can adversely affect not just the developers but also the NMT system's end-users, leading to potential biases or inaccuracies \cite{gianni2022governance}, reminiscent of societal challenges highlighted in Hosea 4:6 due to ignorance. This scenario advocates for a balanced approach that marries technical skill with ethical mindfulness, echoing the societal virtues of wisdom and learning. The involvement of Institutional Review Boards (IRB) in emphasizing data privacy, informed consent, and risk management aligns with the Biblical appreciation for wise counsel (Proverbs 12:15), promoting the development of responsible and reliable NMT systems.

What's more, linguists' feedback on NMT system output can be invaluable to developers who continually seek to refine and enhance the capabilities of these systems. It is paramount, however, that developers maintain an open-minded approach to iteration\cite{Haroutunian2022EthicalCF}. Navigating challenges such as handling out-of-vocabulary words, adapting to language evolution, and tailoring NMT systems to specific tasks requires a commitment to ongoing improvement and flexibility. Related studies also emphasise this approach by suggesting various feedback incorporation approaches such as Human-in-the-loop learning context based education systems, re-read and feedback mechanisms and mixed reality learning.

Finally, in this paper the aurthors coil the phrase 'Blame is to human, not the Model'. It aptly encapsulates the narrative that AI models, in themselves, do not possess ethical agency; rather, the responsibility lies with the developers. A study evaluating the performance of GPT-4 in responding to complex medical ethical vignettes demonstrated that while the model could identify and articulate key ethical issues, it struggled with the nuanced aspects of ethical dilemmas and misapplied certain moral principles \cite{balas2023exploring}. This highlights the limitations of models in comprehending real-world ethical complexities. Nonetheless, this recognition does not diminish the ongoing efforts to establish frameworks for adhering to ethical guidelines in the deployment of AI systems within organizations. Biblically, this concept echoes the narrative in Genesis 1:28, where God grants man dominion over the earth, entrusting humans with stewardship and responsibility over all creation. This scripture has often been interpreted as bestowing upon humanity not just authority, but also the accountability for their actions and decisions, a principle that resonates with the modern ethical discourse surrounding AI development and use.

\subsection{Recommendations}
Developers must prioritize creating NMT systems that are not only efficient but also ethically responsible, transparent, and fair. As NMT contributes significantly to the AI ecosystem, its role in advancing towards Artificial General Intelligence (AGI) emphasizes the need for a strong ethical foundation. This is particularly relevant as AI legislation evolves worldwide. The academic world, merging with AI progress, encounters distinct ethical and regulatory challenges, highlighting the necessity for precise guidelines for ethical AI application in educational technologies\cite{Caccavale2022ToBF}. Engaging actively with legal standards and integrating AI in education demands a commitment to ethical practices, akin to the wisdom in "Let all things be done decently and in order" (1 Corinthians 14:40), ensuring AI serves to enhance, not undermine, educational integrity and inclusivity.

\section{Conclusion}
As reliance on AI and neural network-based tools like NMT systems grows, it's imperative for the next generation of developers to ensure these models are ethically sound and uphold fairness. This begins with comprehensive education during their training, where they learn to navigate ethical challenges in AI development. This paper aims to inspire both new and seasoned developers to view their work through an ethical lens and understand their societal impact. Reflecting on the notion that individuals have roles and purposes as highlighted in spiritual texts, developers are encouraged to see their contributions to AI and NMT not just as technical tasks but as fulfilling a broader societal responsibility, thereby upholding ethical standards and contributing positively to society.

{
\bibliographystyle{ieeetr}
\bibliography{references}
\vspace{12pt}
}

\end{document}